# Mature GAIL: Imitation Learning for Low-level and High-dimensional Input using Global Encoder and Cost Transformation

**Wonsup Shin, Hyolim Kang, Sunghoon Hong**
Yonsei University, Seoul, Korea.
{wsshin2013, simcity429, shhong01 }@yonsei.ac.kr

## Abstract

Recently, GAIL framework and various variants have shown remarkable possibilities for solving practical MDP problems. However, detailed researches of low-level, and high-dimensional state input in this framework, such as image sequences, has not been conducted. Furthermore, the cost function learned in the traditional GAIL framework only lies on a negative range, acting as a non-penalized reward and making the agent difficult to learn the optimal policy. In this paper, we propose a new algorithm based on the GAIL framework that includes a global encoder and the reward penalization mechanism. The global encoder solves two issues that arise when applying GAIL framework to high-dimensional image state. Also, it is shown that the penalization mechanism provides more adequate reward to the agent, resulting in stable performance improvement. Our approach's potential can be backed up by the fact that it is generally applicable to variants of GAIL framework. We conducted in-depth experiments by applying our methods to various variants of the GAIL framework. And, the results proved that our method significantly improves the performances when it comes to low-level and high-dimensional tasks.

## Introduction

Many real-world problems can be represented as Markov decision process (MDP). In addition, advances in storage technology have made it possible to store expert trajectory data on problems. In this context, imitation learning (IL), a method that can directly imitate expert behavior, has attracted much attention as a method for efficient reinforcement learning (RL) agent. Among them, generative adversarial imitation learning (GAIL) approach is showing tremendous performance over traditional IL approaches (Ho and Ermon, 2016). It was also verified that GAIL frameworks work well on high-dimensional tasks consisting of 376 sensor information. Moreover, various follow-up studies have been proposed to construct a hierarchical policy and enhance the balance of learning (Li et al., 2017; Sharma et al, 2018; Peng et al. 2018).

However, real-world problems often provide only low-level and high-dimensional state inputs such as image sequences. For example, an autonomous driving car task (Sallab et al., 2017) and a game task such as Atari (et al. 2015, Hessel et al. 2018) or Minecraft (Oh et al., 2016) comes with a raw image sequence an input. Elements of these inputs do not directly map meaningful information by themselves and the input itself consists of over the thousands of dimensions. So, the agent should be able to extract meaningful information from these inputs.

When dealing with these inputs in the GAIL framework, the issue arises where state dimensions dominate action dimensions. The discriminator of the GAIL framework is a multi-modal model that receives state-action pairs as input. According to multi-modal studies, differences in input dimensions lead to imbalance of importance (Atrey et al. 2010). In order to solve this issue, a method of configuring an additional encoder for state input in the discriminator is mainly used (Atrey et al. 2010). But, in the GAIL framework, there is also a balancing issue between generator and discriminator, since learning in the generative adversarial learning setting is inherently unstable (Goodfellow et al., 2015). That is, in the above case, the learning instability can be aggravated because each network looks at the state from different perspectives.

In this paper, we propose a novel extension model of GAIL framework that can solve low-level and high-dimensional input issues. The proposed model has a global encoder structure in which the generator and discriminator share the encoder of state. This improves stability by inducing two components to give the same perspective on the state. A detailed description of how to learn a stable global encoder is presented in section 3. We also propose a simple but powerful reward shaping mechanism in GAIL framework. All rewards earned through the existing GAIL framework are positive. According to (Sutton and Barto, 1998), this reward function makes it difficult for the agent to reach optimal policy. The proposed reward shaping mechanism, reward penalization, adjusts the range of re-wards to include negative numbers, providing agents with more useful rewards. We call this extension model as MAIL, mature adversarial imitation learning. Finally, the proposed approach can be used generally for the GAIL framework. Therefore, we proved its usefulness by applying the variants of GAIL framework.

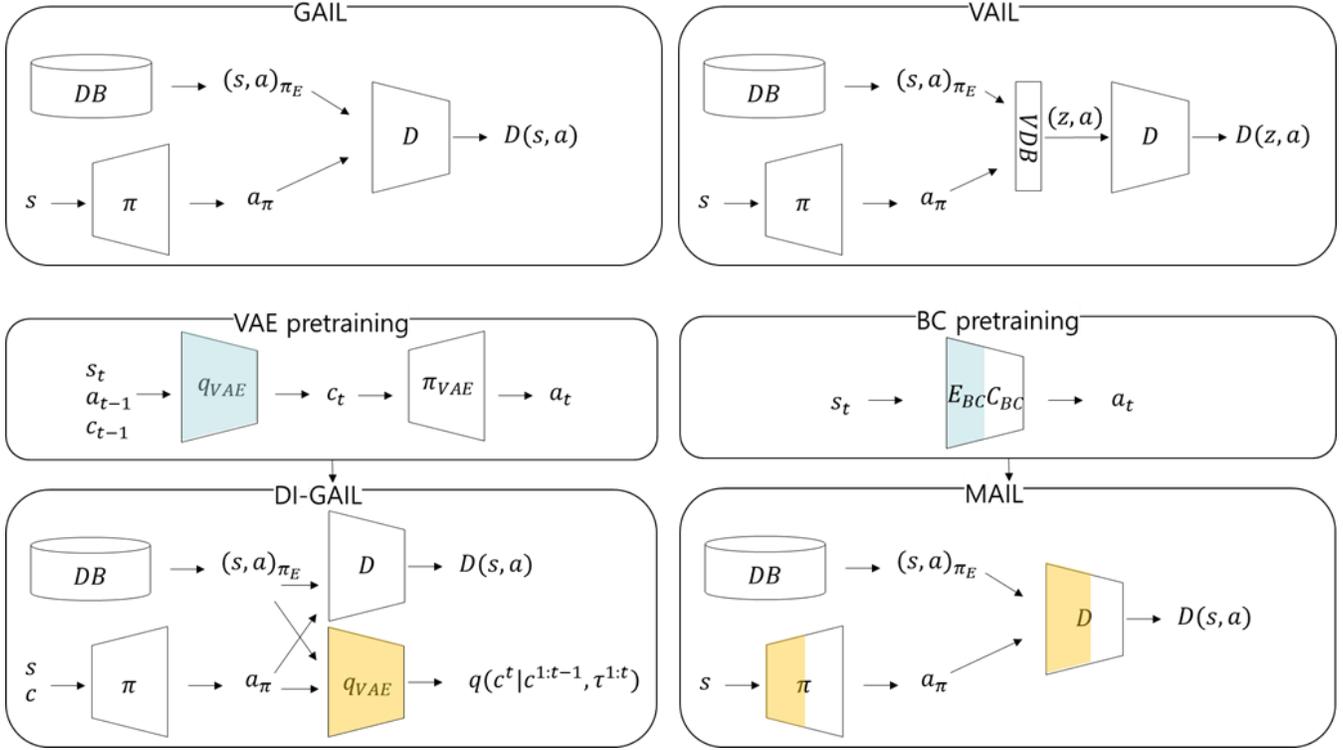

**Figure 1.** Schematic diagrams of GAIL, VAIL, DI-GAIL and MAIL

## Background

**Imitation Learning**

Although reinforcement learning can solve MDP problems, there are lots of cases that the reinforcement signal $r$, which is necessary to run reinforcement learning, is not provided. For this cases, imitation learning tries to yield best policy for the task only by using provided expert trajectories. There are two main approaches of IL. The first is behavioral cloning (BC), which tries to yield a best policy by adopting supervised learning over the expert state-action pairs (Pomerleau, 1991). The second is inverse reinforcement learning (IRL). It tries to find optimal cost function $c$ which derives best reward schemes that can explain the given expert trajectories (Andrew and Russell, 2000; Ziebart et al, 2008; Ziebart et al, 2010). Equation 1 shows typical object function of IRL.

$$maximize_{c \in C}(min_{\pi \in \Pi} - H(\pi) + \mathbf{E}_\pi[c(s,a)]) \quad (1)$$
$$- \mathbb{E}_{\pi_E}[c(s,a)]$$

Where $H(\pi) \equiv E_\pi[-\log \pi(a|s)]$ denotes the $\gamma$-discounted entropy of the policy $\pi$, and $\pi_E$ denotes the expert policy that is given as sampled trajectories in practice.

**Generative Adversarial Imitation Learning**

Conventional IRL approaches require additional RL step over the reward scheme derived from the IRL to get the best policy for the given task. However, inspired by GAN, GAIL derive the best policy directly from given expert trajectories (Ho et al. 2016). The formal GAIL objective is following.

$$min_\pi max_{D_{s \sim S, a \sim A} \in (0,1)} \mathbb{E}_\pi[\log D(s,a)] \quad (2)$$
$$+ \mathbb{E}_{\pi_E}[\log(1 - D(s,a))]$$

Where $D$ denotes the discriminator, which tries to distinguish state-action pairs from the trajectories generated by $\pi$ or $\pi_E$. Theoretically, it is proved that optimizing equation 1 includes both IRL and RL step.

**Variants of Generative Adversarial Imitation Learning**

Variational adversarial imitation learning (VAIL) is a method of adjusting the balance between generator and discriminator by giving a constraint to the discriminator using a variational encoder called a variational discriminative bottleneck (VDB). This method adds a term to the object function that maximizes the mutual information between E(z|x) and r(z) so that the discriminator can produce a significant reward.

DI-GAIL (Directed-Info GAIL) is a model that agent can learn hierarchical policy without knowledge of option. In order to learn option, they use directed information (Kramer, 1998) as a measure to map option to latent variable $c$. Therefore, they add a term that maximizes the directed information between c and trajectory $\tau = (s_1, a_1, s_2, a_2, \ldots s_{t-1})$ In order to approximate the distribution of c necessary for the use of the objective function, the approximate function q is trained by the pre-training phase and then transferred.

While the above two variants improve the stability based on information theory, the proposed model improves stability through structural modification, so it can be easily combined with the variants without friction.

## Mature Generative Adversarial Imitation Learning

In this section, we present overall structure and detailed description of our approach. Figure 1 shows the schematic diagram of GAIL, VAIL, DI-GAIL, and MAIL. The light blue part of the component means that it is transferred to the next phase and the orange part means the part that has been transferred.

**Global Encoder**

Our idea is simple. It has an encoder for the state shared by the RL agent and discriminator. This alleviates the difference in dimension between state and action in the discriminator and does not compromise the balance between the agent and the discriminator. Since global encoders that are involved in all components have a key impact on performance, we put two learning phases for stable learning of global encoders. In the first phase, we train the actor $\pi_{BC} = \{E_{BC}, C_{BC}\}$ using BC algorithm. Where $E_{BC}$ is encoder part of actor and $C_{BC}$ is classifier part of actor. After that, the trained $E_{BC}$ is transferred to the global encoder and fixed, and the MAIL is learned in the same manner as the GAIL. Because BC does supervised learning, it can model the expert trajectory most robustly. The object function of MAIL is defined as equation 3.

$$min_\pi max_{D_{s \sim S, a \sim A} \in (0,1)} \mathbb{E}_\pi[\log D(E_{BC}(s), a)] + \mathbb{E}_{\pi_E}[\log(1 - D(E_{BC}(s), a))] \quad (3)$$

**Reward Penalization**

From the agent's point of view, $min_\pi E_\pi[\log D(s, a)]$ part of equation 2 can be reinterpreted as a reward sign $R(s, a) = -\log D(s, a)$. Note that the domain of $D(s, a)$ is [0,1] and the equilibrium is formed when $D(s, a) = 0.5$ for all the (s, a) pairs. If $R(s, a) = -\log D(s, a)$, our agent will get positive reward sign for every step, even though the agent did not learn at all. To solve it, we suggest new reward sign $R(s, a) = -\log(D(s, a) + 0.5)$ that satisfies $R(s, a) = 0$ when $D(s, a) = 0.5$. Not only this reward transformation results in stable performance near the equilibrium, it also provides richer reward sign significant to the agent when it comes to adequate learning because now the transformed reward function has a negative range.

Finally, we summarize the learning algorithms of MAIL and DI-MAIL in Algorithm 1 and 2, respectively.

---

**Algorithm 1. Mature GAIL (MAIL)**

**Phase 1: Pre-training encoder step**
Input: expert trajectories $\tau_E \sim \pi_E$, initial global encoder, actor network parameters $\eta_0, \alpha_0$

**for** i = 0, 1, 2, ⋯, n:

    1. Sample τ from $\tau_E$
    2. Update the $\eta_i \rightarrow \eta_{i+1}, \alpha_i \rightarrow \alpha_{i+1}$,
       with minimize{L = $-\log \pi_{BC}(s)_i$}
Output: global encoder parameter $\eta_n$

**Phase 2: Main step**
Input: expert trajectories $\tau_E \sim \pi_E$, initial actor, critic and discriminator network parameters $\alpha_0, \beta_0, \delta_0$, and trained global encoder parameter $\eta_n$ from phase 1.
  1. Load $\eta_n$ to the global encoder and fix
  2. Learn under GAIL

---

**Algorithm 2. Mature DI-GAIL (DI-MAIL)**

**Phase 1: Pre-training encoder step**
Input: expert trajectories $\tau_E \sim \pi_E$, initial global encoder, actor network parameters $\eta_0, \alpha_0$

**for** i = 0, 1, 2, ⋯, n:

    1. Sample τ from $\tau_E$
    2. Update the $\eta_i \rightarrow \eta_{i+1}, \alpha_i \rightarrow \alpha_{i+1}$,
       with minimize{L = $-\log \pi_{BC}(s)_i$}
Output: global encoder parameter $\eta_n$

**Phase 2: Pre-training posterior step**
Input: expert trajectories $\tau_E \sim \pi_E$, initial actor and posterior network parameters $\alpha_0, \varphi_0$, and trained global encoder parameter $\eta_n$.

**for** i = 0, 1, 2, ⋯, n:

    1. Sample τ from $\tau_E$
    2. Sample $c_i$ from posterior network
    3. Update the $\varphi_i \rightarrow \varphi_{i+1}, \alpha_i \rightarrow \alpha_{i+1}$,
       with minimize{ $L_{VAE}$ loss on (Sharma et al., 2018)}
Output: posterior parameter $\eta_n$

**Phase 3: Main step**
Input: expert trajectories $\tau_E \sim \pi_E$, initial actor, critic and discriminator network parameters $\alpha_0, \beta_0, \delta_0$, and trained global encoder and posterior network parameter $\eta_n, \varphi_m$ from phase 1 and 2.
  1. Load $\eta_n, \varphi_m$ to the global encoder and posterior and fix
  2. Learn under DI-GAIL

Table 1. Results on the navigation task.

| Model | Best score | score | Meets -10 | After meets -10 |
|---|---|---|---|---|
| GAIL | -97.491 | -99.43±0.80 | - | - |
| VAIL | -99.949 | -100.00±0.01 | - | - |
| GAIL_LS | -1.516 | -31.18±29.54 | 28K | -9.10±9.26 |
| VAIL_LS | -1.237 | -25.71±28.28 | 23K | -10.89±13.02 |
| GAIL_GE | -99.939 | -99.97±0.03 | - | - |
| **MAIL** | **1** | **-6.63±15.56** | **13K** | **-3.37±9.26** |
| **MAIL + VDB** | **0.996** | **-3.45±11.97** | **3K** | **-2.39±8.59** |
| DI-GAIL_GE | -92.824 | -96.91±2.05 | - | - |
| **DI-MAIL** | **1** | **-6.00±17.79** | **3K** | **-4.12±13.32** |
| **DI-MAIL + VDB** | **0.995** | **-4.84±14.28** | **3K** | **-3.18±9.78** |

## Experiments

We demonstrate the effectiveness of our method on a hierarchical navigation task. In addition, we also investigate various methods for reward penalization on the LunarLander-v2 environment. We will show that (1) global encoder is able to learn meaningful representation of raw image pixel input, (2) reward penalization has remarkable effect to performance and (3) our method is able to be applied to GAIL framework-based variants.

The source codes of our experiments can be seen at https://github.com/sunghoonhong/Mature-GAIL

### Environments

To validate our proposed approach, we choose hierarchical navigation task in grid world environment, which consists of a 7 x 7 grid with four rooms connected via bottleneck passage. Each grid is represented by 4 x 4 pixel with RGB formulation. So, we got 32 x 32 x 4 size state input. The agent spawns at a random grid and its goal is to reach a key, then reach a car. Both key and car spawn at a random grid in top left room and bottom right room each. The reward is given as much as shortest distance when the agent achieves the goal, otherwise -1 for each timestep. We utilize about 1M state-action pairs generated by shortest path algorithm as expert demonstration.

We also experiment in LunarLander-v2 environment, provided in OpenAI Gym (Brockman et al., 2016). The agent spawns at the top of the screen and its goal is to land on the landing pad. The action can be firing main, left or right engine or doing nothing. The state consists of position, velocity, angle, angle velocity and contact of legs. The reward is given for leg ground contact or landing on landing pad. On the other hand, the penalty is given for firing engine or crashing. We use about 10K state-action pairs generated by the agent trained using PPO (Schulman et al., 2017) algorithm as expert demonstration.

On the hierarchical navigation task, we conduct three experiments: demonstrating that our method is able to learn in raw image pixel space, analyzing encoded states, and applying our method to GAIL framework-based variants on various settings, including DI-GAIL for hierarchical learning. Furthermore, we analyze several ways of reward penalization on LunarLander-v2 environment.

### Variants of MAIL

We combine our proposed approach, MAIL, to VAIL and Directed-Info GAIL. In addition, we also apply it to GAIL without global encoder for demonstrating that reward penalization is effective even in another experiment. For implementation detail, we use PPO algorithm for training agents rather than TRPO (Schulman et al., 2015).

### Performance results

We apply each component step by step to various variations on the navigation task and show the quantitative evaluation in terms of performance and learning stability. The result is calculated by the score which is a mean return over 1000 episodes. We assume that the agent has learned enough if the score meets -10. As can be seen in Table 1, GAIL framework-based variants applied our method show superior performance rather than naïve methods without ours. Firstly, naïve methods cannot solve the task at all. And it seems that the agents only with global encoder improve very little bit, but still cannot solve either. On the other hand, the agents with only with reward penalization show much better performance but still cannot completely solve either. As a result, the agents with our method meet score 1 which is the upper bound and show stable performance

**Table 2. Comparison of reward penalization schemes on navigation task.**

| Scheme | Score |
|---|---|
| Log | -99.97±0.03 |
| Log scaled | -97.43±2.48 |
| Log shift | -6.63±15.56 |
| Linear | -5.02±14.18 |
| Tan | -6.50±17.88 |

**Table 3. Comparison of reward penalization schemes on LunarLander-v2 environment.**

| Scheme | Score | Solved (%) | Collapsed (%) |
|---|---|---|---|
| Log | 91.84±86.53 | 0 | 0 |
| Log scaled | 92.46±87.77 | 0 | 10 |
| Log shift | 135.91±99.24 | 20 | 10 |
| Linear | 146.40±97.11 | 30 | 10 |
| Tan | 187.90±91.62 | 40 | 0 |

**Analysis on global encoder**

In Figure 2, we demonstrate that (1) loading the weights of global encoder from behavior cloning and (2) fixing the weight of global encoder during learning are necessary. To show these we experiment in other strategies. We experiment for both MAIL and MAIL + VAIL. 'F' means fixing the weights of global encoder, 'L' means loading the weights from Behavior Cloning pre-training. One does not load the weights from Behavior Cloning pre-training but randomly initialize and train. The other loads the weights from the pre-training and does not fix the weights during training. We skip the case which randomly initializes and fix the weights. Only with our proposed strategy the agent is trained properly. On the other hand, agents with other strategies fail to learn to get enough score, and most of them collapse as learning progresses. For the reason, we suspect that loading and fixing the weights reduce instability in GAIL framework which is inherently fluctuating.

Furthermore, we analyze how the encoded states are distributed in Figure 3. For the states at the bottom and right side, in case of the agent hasn't taken key yet, the encoder gives a focus on the position of key rather than the position of car. For the states at the top, in case of the agent has already taken the key, it seems that encoder gives a focus on the distance between the agent and the car. Additionally, while the left-side state of the top and the left-side state of the bottom has the same distance between the agent and car, the distance of encoded states is huge. The only difference between two states is whether the agent has taken the key or not. These show that the encoder works in terms of informative encoding and give a focus on which is important to the agent.

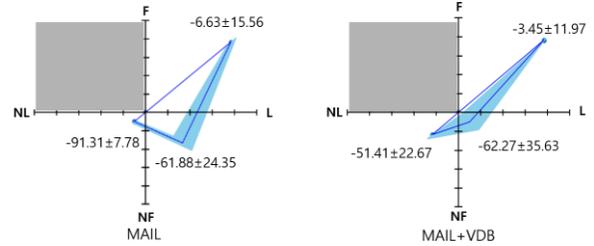

**Figure 2. Comparison of the training strategy for global encoder.**

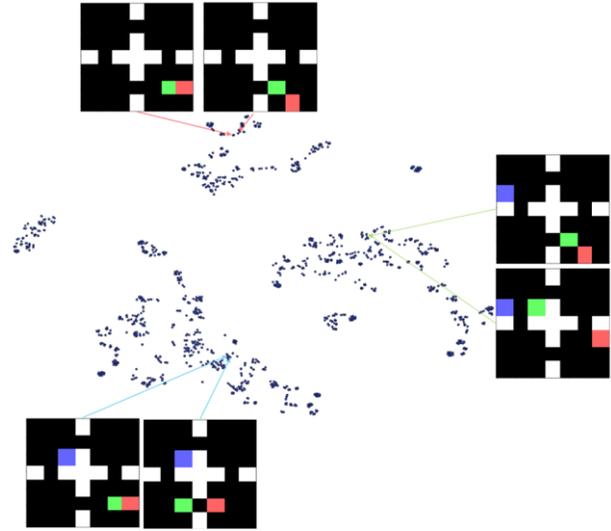

**Figure 3. Visualization of encoded state using t-SNE.**

**Analysis on reward penalization**

For comparison of several schemes, we experiment using MAIL with 5 reward schemes on two environments. For LunarLander-v2 environment, we assume that it is solved if the score is over 200 and collapsed if the score is under 0 after learning.

We investigate several reward penalization schemes on two environments. For a baseline, we use the original log reward which is always positive, and scaled log reward which is divided by 10 so that the reward is bounded in smaller range. Then we compare shifted log reward, linear reward and tangent reward which is positive in [0, 0.5) and negative in (0.5, 1]. We used $\tan(0.5 - D(s, a))$, $0.5 - D(s, a)$ for each tangent reward and linear reward.

As can be seen in Table 2,3, it is obvious that the agents trained under reward penalization show remarkably high performance rather than non-penalization. On the other hand, there is no superior scheme among three penalization schemes. On the navigation task, linear scheme shows the best result, but on LunarLander-v2 environment, tangent scheme seems that the most effective scheme. As a result, we demonstrate that reward penalization significantly improves the performance, and the choice of scheme can be a hyperparameter.

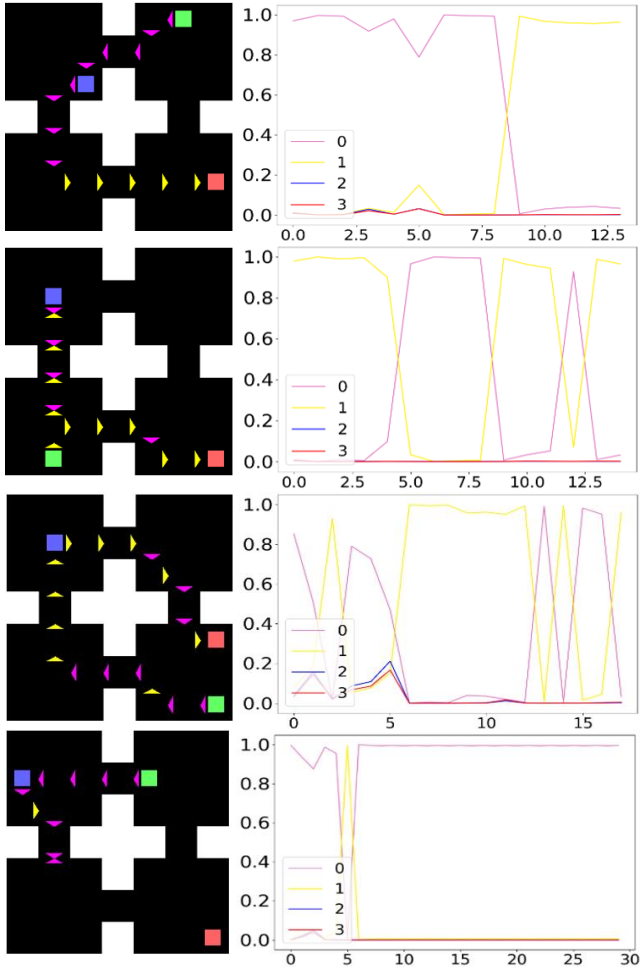

**Figure 4. Trajectories and predictions of each code for DI-MAIL on navigation task.**

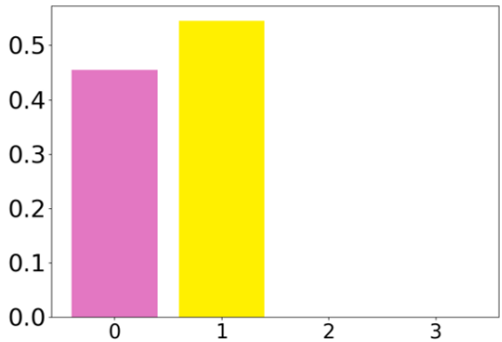

**Figure 5. Proportion of codes during total episodes.**

**Analysis on latent code**

We also apply our method to DI-GAIL which is able to learn hierarchical policy based on option framework. We set the latent code as four different categorical variables. In Figure 4, the arrow means the action by its direction and the code by its color. According to Figure 5, only two code variables are used which are unsupervisedly learned from pre-training. The agent appropriately uses two codes as episode proceeds. It seems that each code corresponds to different traversal strategy. The code 0 denoted as pink color tends to direct the agent to traverse along left and downward, while the code 1 denoted as yellow color tends to direct the agent to traverse along right and upward. The last trajectory in Figure 4 shows that the agent which is trained using naïve method fails to learn. While the code in $8^{th}$ timestep tells the agent to move downward, it chooses to move upward. It means that pre-trained distribution of code properly provides the agent to choose correct actions, even there is some possibilities of learning failure of the agent. From the above results, we can say it is possible that the DI-MAIL agent is able to learn consistent and meaningful latent code variables in unsupervised method and solve the problem which has hierarchy.

**Discussion**

Adopting the pretrained encoder showed meaningful performance improvement. It seems that using pre-trained global encoder through behavior cloning mitigates inherent instability of GAN framework. However, reconstruction pre-trained used in the World Model (Ha et al, 2018) doesn't work in our experiment. Further study to find better structure or pre-training method for global encoder is needed. For instance, adopting transfer learning to the global encoder can be a feasible attempt.

Moreover, it was revealed that penalizing reward played significant role when it comes to performance improvement of imitation learning task. Due to its ease of implementation and potential of general application, it can be considered as meaningful contribution.

Furthermore, since there are many real-world problems which have complicated hierarchical structure and high-dimensional state space, especially raw image, we expect high potential in our method in that it is able to learn hierarchical policy from raw image pixel inputs.

## Conclusion

We proposed MAIL, a novel GAIL framework that is adaptable to tasks that use low-level and high-dimensional inputs. The key ideas are the global encoder and reward penalization mechanism. Also, the proposed method is generally available for GAIL framework. As a result of in-depth experiments, the proposed method outperforms the existing methods, and further experiments by idea demonstrate the usefulness of the proposed method.